\documentclass[conference]{IEEEtran}
\IEEEoverridecommandlockouts
\usepackage{cite}
\usepackage{amsmath,amssymb,amsfonts}
\usepackage{algorithm}
\usepackage{graphicx}
\usepackage{textcomp}
\usepackage{xcolor}
\usepackage{comment}
\usepackage{float}
\usepackage{algpseudocode}
\usepackage{enumitem}
\usepackage[para,flushleft]{threeparttable}
\usepackage[export]{adjustbox}[2011/08/13]
\usepackage{caption}
\usepackage{bbding}
\usepackage{makecell}
\usepackage{arydshln}
\usepackage{cuted}
\usepackage{subcaption}
\usepackage{svg}
\usepackage{bm}
\usepackage{hyperref}
\usepackage{titlesec}

\def\BibTeX{{\rm B\kern-.05em{\sc i\kern-.025em b}\kern-.08em
    T\kern-.1667em\lower.7ex\hbox{E}\kern-.125emX}}
\begin{document}

\title{From Detection to Action Recognition: An Edge-Based Pipeline for Robot Human Perception}

\author{\IEEEauthorblockN{Petros Toupas, Georgios Tsamis, Dimitrios Giakoumis, Konstantinos Votis, Dimitrios Tzovaras}
\IEEEauthorblockA{\textit{Information Technologies Institute} \\
\textit{Centre for Research \& Technology Hellas}\\
Thessaloniki, Greece \\
\{ptoupas,gtsamis,akargakos,dgiakoum,kvotis,Dimitrios.Tzovaras\}@iti.gr}
}

\maketitle

\begin{abstract}
Mobile service robots are proving to be increasingly effective in a range of applications, such as healthcare, monitoring Activities of Daily Living (ADL), and facilitating Ambient Assisted Living (AAL). These robots heavily rely on Human Action Recognition (HAR) to interpret human actions and intentions. However, for HAR to function effectively on service robots, it requires prior knowledge of human presence (human detection) and identification of individuals to monitor (human tracking). In this work, we propose an end-to-end pipeline that encompasses the entire process, starting from human detection and tracking, leading to action recognition. The pipeline is designed to operate in near real-time while ensuring all stages of processing are performed on the edge, reducing the need for centralised computation. To identify the most suitable models for our mobile robot, we conducted a series of experiments comparing state-of-the-art solutions based on both their detection performance and efficiency. To evaluate the effectiveness of our proposed pipeline, we proposed a dataset comprising daily household activities. By presenting our findings and analysing the results, we demonstrate the efficacy of our approach in enabling mobile robots to understand and respond to human behaviour in real-world scenarios relying mainly on the data from their RGB cameras.
\end{abstract}


\section{Introduction} \label{intro}

The integration of robots into human lives is increasingly common as people become more comfortable interacting with them. In healthcare, autonomous mobile service robots show great promise in supporting caregivers and family members responsible for patients. Moreover, ambient assisted living benefits from robotic companions that monitor the well-being of elderly or impaired individuals in various settings such as homes, nursing homes, and hospitals. To enable robots to comprehend and respond to human behaviour, human action recognition systems play a vital role. These systems allow robots to deduce information about a person's state, behaviour, and intentions, ranging from providing emergency assistance to understanding daily activities. Nonetheless, before implementing such capabilities on service robots, certain preceding steps need to be addressed. In practice, HAR represents the final stage of a larger pipeline, as discussed in detail in Section \ref{sec:pipeline}. The pipeline commences with user recognition and identification among multiple individuals. A tracking algorithm is then employed to follow the user within the robot's operating space. Once these prerequisites are fulfilled, the action recognition algorithm accurately gathers user information \cite{Hassan2017}. In Autonomous Mobile Robot (AMR) applications, HAR is crucial as the robot's behaviour depends on the activities of its human counterpart. For instance, a vision-based system should continuously recognise human actions in a live streaming scenario without prior knowledge of when actions start or end.

Deploying human action recognition systems on mobile robots requires careful and precise design of their pipelines, along with the selection of lightweight algorithms for each stage. However, there are several challenges that arise when deploying vision-based HAR algorithms on mobile robots, as discussed in \cite{Jegham2020b, Mostafa2023}. These challenges include variations in point of view, illumination, shadows, and scale due to the different positions the robot may take while operating as discussed by Ahmad et al. \cite{Ahmad2015UsingDC}. Near real-time, on-device performance is a crucial requirement for such systems, enabling the robot to respond and plan accordingly to human actions. Furthermore, optimising resource utilisation is essential to avoid overburdening the system. Offloading part of the processing to a server or the cloud may restrict the system's performance due to limited bandwidth or other communication barriers.

This paper introduces an edge-based end-to-end human action recognition pipeline designed for mobile service robots. To achieve near real-time performance, we employ efficient and lightweight algorithms such as OpenPose and X3D at each stage of the pipeline. Extensive research and analysis have been conducted to select the most suitable HAR model, with a focus on minimising resource utilisation. To evaluate the effectiveness of our approach, we utilise a newly introduced dataset that comprises seven distinct actions. This dataset combines data from various publicly available datasets along with newly captured data from a demonstration infrastructure for rapid prototyping and novel technologies at our premises. The main contributions of this paper can be summarised as follows:

\begin{itemize}
    \item An end-to-end solution for recognising human actions via a mobile robot in near real-time with all processing in each stage performed on the edge utilising a 3D-based user tracking algorithm and an overlapping sliding window pre-processing technique for higher-quality HAR predictions.
    \item The introduction of a dataset on daily activities that combines data from various public datasets with newly generated data from a Smart House environment dedicated to rapid prototyping and emerging technologies.
    \item A thorough evaluation and comparison of state-of-the-art human action recognition models on the proposed dataset, with a particular focus on resource utilisation and near real-time, on-device performance.
\end{itemize}

\section{Related Work} \label{sec:background}

In their work, Kumar and Sukavanam \cite{8441712} delve into the critical analysis of human activities, focusing on spatial and temporal dynamics. Their study concentrates on skeleton-based human activity recognition, proposing a novel motion trajectory computation method that utilizes Fourier temporal features derived from interpolated skeleton joints. Yucer and Akgul \cite{Yucer20183DHA} proposed an approach defining the 3D human action recognition as a Deep Metric Learning (DML) challenge, wherein a similarity metric is acquired between distinct 3D joint sequence data via deep learning techniques. The proposed DML network adopts a Siamese-LSTM (S-LSTM) architecture, featuring two parallel networks with shared parameters.

In recent years, there have been several attempts to transition from assessing the performance of HAR models on static datasets to evaluating their performance in real-world scenarios when deployed on edge devices such as mobile robots. Xia et al. \cite{7045908} proposed a framework that adopts an RGBD camera to explore human-robot activities and interactions from the viewpoint of the robot. They captured two multi-modal datasets comprising RGB, depth, and skeleton data, and evaluated their system using these datasets. In their work, Rezazadegan \textit{et al.} \cite{Rezazadegan2017ActionRF} introduced a novel approach for generating action region proposals that are robust to camera motion. They utilised convolutional neural networks (CNNs) to jointly detect and recognise human activities. They evaluated their HAR system on both existing public datasets and newly introduced ones. J. Lee and B. Ahn \cite{s20102886} utilised two open-source libraries, namely OpenPose \cite{8765346} and 3D-baseline, to extract skeleton joints from RGB images using convolutional neural networks to classify human actions. Their system was deployed on a robotic platform equipped with an NVIDIA JETSON XAVIER and evaluated using the NTU-RGBD benchmark \cite{shahroudy2016ntu}.

S. Hoshino and K. Niimura \cite{10.1007/978-3-030-01370-7_40} proposed a robot vision system that employs optical flow to describe original images and classify human action. The system consists of two convolutional neural network classifiers for image inputs as well as a novel detector for extracting partial images of the target human. To facilitate camera movement, a set of optical flow modifications were proposed and implemented. Hsieh et al. \cite{9397242} presented an online Human Action Recognition system that detects humans using OpenPose and tracks them using DeepSort \cite{Wojke2018deep} before performing action recognition. An LSTM-based classifier utilising RGB, optical flow, and skeleton features to classify actions was proposed. The method was evaluated on the CVIU Moving Camera Human Action dataset, demonstrating promising results. However, the combination of models used at each stage may limit edge deployment and real-tiem performance. Wang et al. \cite{9397242} proposed an architecture for human action recognition at the AI's edge. They are based on OpenPose and DeepSort for human detection and tracking, and on bidirectional long-short-term memory (DLBiLSTM) for human action recognition, with input from the DenseNet121 feature extractor. To increase system efficiency, they proposed two optimisations: Robot Operating System (ROS) \cite{ros_framework} distributed computation and TensorRT structure optimisation, which allowed them to perform and execute their pipeline entirely on their autonomous mobile robot.

The current solutions frequently fail to fulfil or overlook the near real-time demands and constraints posed by robotic platforms, since they evaluate their methods on hardware unsuitable for mobile robots, such as the NVIDIA GTX 1080 used in previous works, \cite{s20102886}, \cite{10.1007/978-3-030-01370-7_40}, and \cite{9397242}. In contrast, our proposed pipeline focuses primarily on efficiency and deployability directly on the edge, specifically on the mobile robot, without the need for external hardware offloading. Moreover, we carefully consider and compare against state-of-the-art models, including X3D \cite{Feichtenhofer2020}, Swin-Transformer \cite{9710580}, and TimeSformer \cite{Bertasius2021IsSA}, while building our HAR system.

\section{Pipeline Architecture} \label{sec:pipeline}

\begin{figure}[!hbt]
    \centering
    \includegraphics[width=1.0\linewidth]{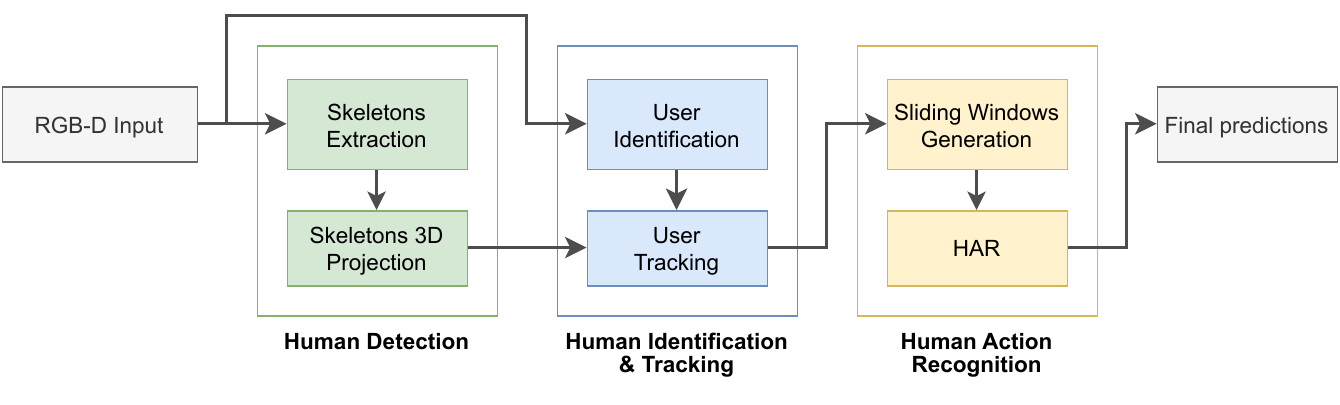}
    \caption{The proposed pipeline architecture deployed end-to-end on the mobile robotic platform, from human detection and tracking to action recognition.}
    \label{fig:architecture}
\end{figure}

Our proposed pipeline incorporates and combines a number of algorithms at each phase, making the required changes to them as needed to adapt to the robotic system's dynamic and near real-time nature. Figure \ref{fig:architecture} depicts a high-level overview of the pipeline and its components. Human detection, human tracking, and human action recognition are the three basic processes involved in the execution of the robot's pipeline, and each of them is further broken down into a finer description made of sub-modules.

\begin{figure*}[!hbt]
   \begin{subfigure}{0.24\textwidth}
       \includegraphics[width=\linewidth]{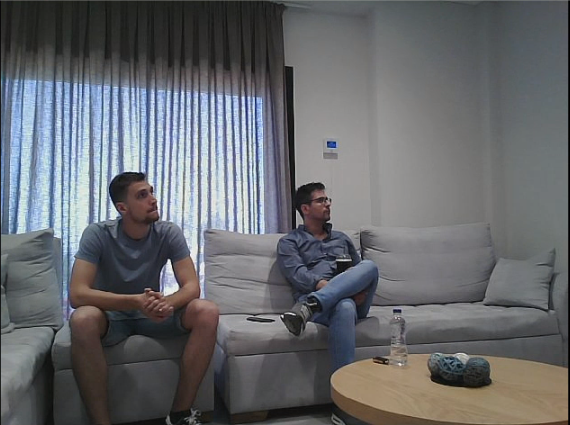}
       \caption{Raw RGB image from the robot's camera}
       \label{fig:rgb_raw}
   \end{subfigure}
\hfill %
   \begin{subfigure}{0.24\textwidth}
       \includegraphics[width=\linewidth]{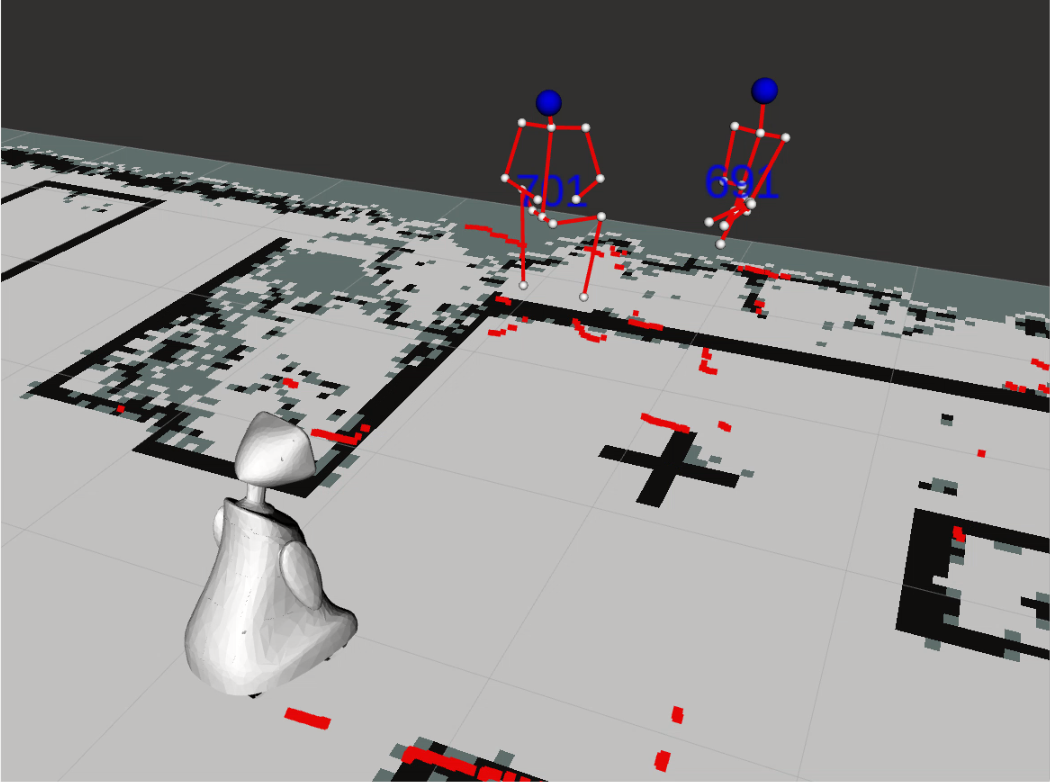}
       \caption{Projection of human skeletons to the 3D space}
       \label{fig:3d_map}
   \end{subfigure}
\hfill %
   \begin{subfigure}{0.24\textwidth}
       \includegraphics[width=\linewidth]{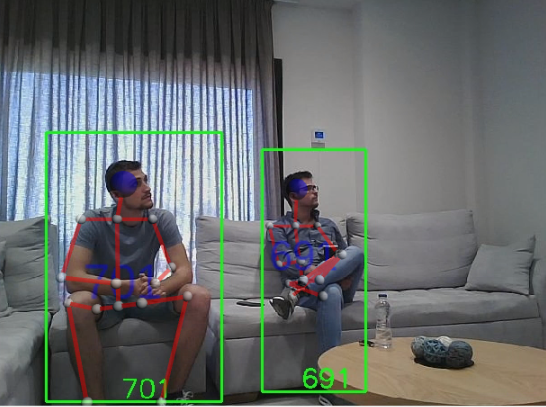}
       \caption{ID assignment to the detected human skeletons}
       \label{fig:rgb_pose_id}
   \end{subfigure}
\hfill %
   \begin{subfigure}{0.24\textwidth}
       \includegraphics[width=\linewidth]{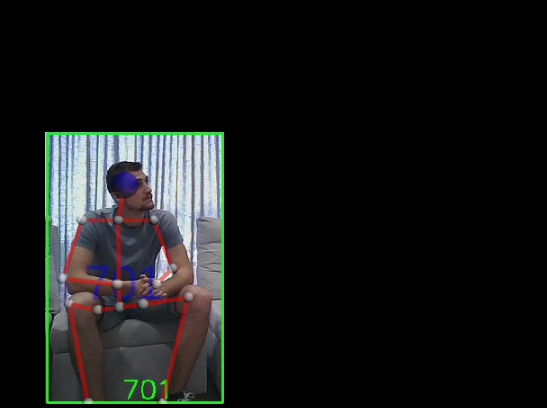}
       \caption{Selection of user skeleton and respective bbox for input to HAR}
       \label{fig:rgb_pose_id_cropped}
   \end{subfigure}

   \caption{The steps of the pipeline from the perspective of the robot Starting with capturing the raw RGB image, detecting the skeletons of the people in the scene, projecting them to 3D space, assigning tracking IDs to each individual, and cropping the user's bbox to provide as input to the HAR model.}
   \label{fig:pipeline_rgb}
\end{figure*}

As shown in Figure \ref{fig:pipeline_rgb} the flow begins by capturing a raw RGB image of the scene. Subsequently, the system detects the skeletons of the humans present in the image. These detected skeletons are then projected into a 3D space, providing the necessary input and allowing the tracking algorithm to be executed. Through this algorithm and to keep track of each individual, the system assigns unique tracking IDs. Once the user is identified, the system crops the corresponding bounding box (bbox) around the user, ensuring that only relevant information is provided as input to the HAR model.

\subsection{Human Detection} \label{subsec:detection}

\begin{figure}[!hbt]
    \centering
    \includegraphics[width=0.9\linewidth]{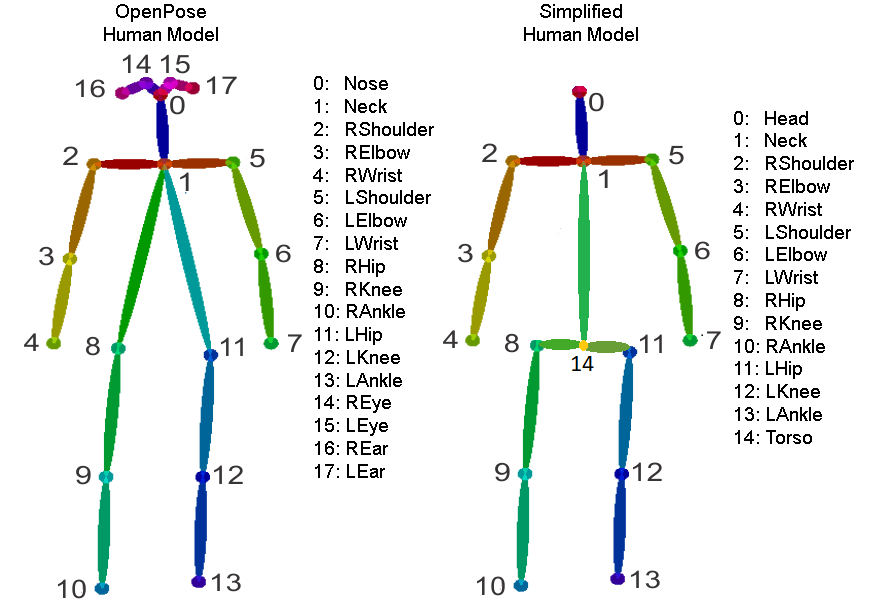}
    \caption{Human skeletal keypoints}
    \label{keypoints}
\end{figure}

The detection of the user in the scene is the initial step in our pipeline. To solve this challenge, we utilise the OpenPose library with input taken from the robot's RGB camera. OpenPose \cite{8765346} provides multi-person 2D pose estimation in order to understand people, their motion, and their presence in a scene. The OpenPose \cite{8765346} model extracts and localises skeletal keypoints of persons using Part Affinity Fields (PAFs). Some of the original model's detected keypoints, like the keypoints for the head, are removed or combined, in order to obtain a more manageable representation of a person in 3D space. 
Figure \ref{keypoints} (left model) depicts the human skeletal keypoints proposed in OpenPose.
Figure \ref{keypoints} (right model) depicts a simplified human skeletal keypoints that we introduced to better match the needs of our pipeline.

%

The final stage of the detection module is to project the humans' skeletal keypoints to 3D space. The required depth information is obtained by the robot's depth camera, which is stored as a 2D image. Back-projecting from a 2D image to a 3D map of the space in which the robot operates requires the camera's intrinsic and extrinsic parameter matrices, as well as the points of interest in the 2D image, as provided by OpenPose. We consider the following pinhole camera equation: $\bm{p = K*R*P}$, where p is the projected point on the frame, $K$ is the camera intrinsic matrix, $R$ is the camera extrinsic matrix, and $P$ is a 3D world coordinate point. In intrinsic matrix $K$, the focal length ($f x$, $f y$) and the principal point ($u 0$, $v 0$) are obtained through camera calibration. Given the $Z$ values from the depth image and the (u, v) points from OpenPose, we solve the pinhole camera equation and obtain the point in the 3D map as follows (assuming (u, v) and (X, Y, Z) are in the camera's coordinate frames, so the extrinsic matrix $R$ is not taken into consideration): %

\begin{equation}
    X=Z\frac{u-u_0}{f_x}, Y=Z\frac{v-v_0}{f_y}
\end{equation}

\subsection{Human Identification \& Tracking} \label{subsec:tracking}

The tracking of the user is the next phase in our pipeline. The identification of the user is a vital step before moving on to tracking. To accomplish that, we make use of a facial recognition algorithm, where a CNN model \cite{7298682} produces 128 features for each of the faces in a frame and compares them to the stored features of already known people. After identifying the user based on their face, we combine this information into 3D skeletons from the previous stage. At this stage, we've recognised the user and can track them in 3D space.

The tracking method we employ is based on information from the previous positions of the user in 3D space and the estimation of distances between these positions and detected skeletons at the current frame. To add memory to the tracking system, we must match the skeleton of the user (identified by face) from the previous frame to one of the detected skeletons in the current frame, and forward all past knowledge. This is important because face recognition will not always be able to detect and match the user's face with a skeleton. It is worth noting that this process is repeated with each new frame captured by the robot and its sensors. When the matching fails, the tracking continues for a few frames as it attempts to recover from an unsuccessful user matching with the skeletons. When the tolerance period expires, the tracking ceases and the robot resumes its search for the user using facial recognition. The proposed tracking algorithm that our system utilises is shown below:

\begin{algorithm}[!hbt]
\caption{Human Tracking in 3D}\label{alg:human_tracking}
\begin{algorithmic}
\State $userSkel \gets prevKnownSkel $ \Comment{Known user skeleton}
\State $diameter \gets 1$ \Comment{Meters}
\For{$sk$ in $skeletons$}
\State $curDistance \gets Distance(sk,userSkel)$ \Comment{Euclidean}
\If{$curDistance < minDistance$}
    \State $minDistance \gets curDistance$
    \State $matchedSkel \gets sk$
\EndIf
\EndFor
\If{$minDistance \leq diameter$}
    \State $userSkel \gets matchedSkel$
\Else
    \If{$timesUntracked > 5$}
        \State $userSkel \gets LOST$
    \Else
        \State $timesUntracked \gets timesUntracked + 1$
        \State $userSkel \gets UNKNOWN$ \Comment{Tolerance}
    \EndIf    
\EndIf
\end{algorithmic}
\end{algorithm}

\subsection{Human Action Recognition} \label{subsec:har}

\begin{figure*}[!hbt]
    \centering
    \includegraphics[width=0.75\textwidth]{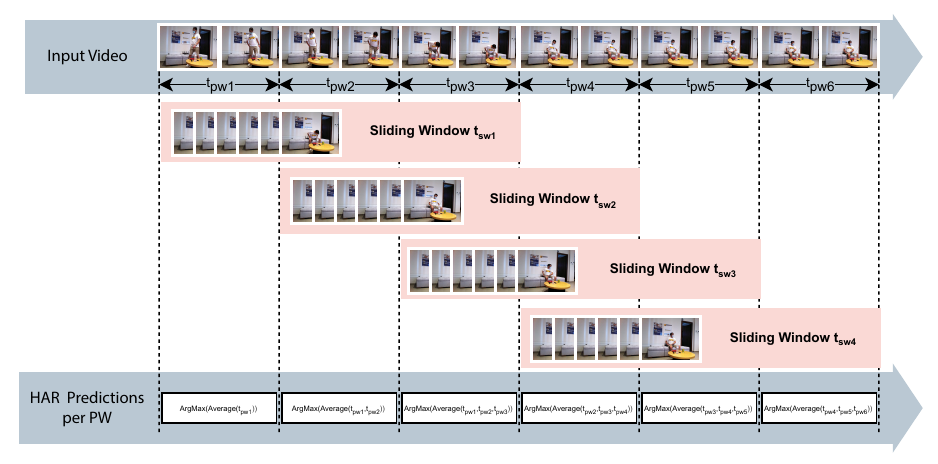}
    \caption{An overview of the proposed sliding window overlap methodology. The input RGB feed is displayed at the top of the figure, organised into time windows of equal duration ($t_pw$). The sliding windows ($t_sw$) span across multiple $t_pw$.}
    \label{fig:har_sliding_window}
\end{figure*}

The final stage in our pipeline is dedicated to recognising human actions based on an input video stream. This data can be leveraged to analyse human behaviour or predict the robot's next state, such as alerting a caregiver about an extended period of the user lying down. Considering the limited processing capabilities of edge devices, it becomes essential to employ efficient and lightweight algorithms for these applications to achieve near real-time performance.

Transitioning from fixed-duration videos, as seen in HAR benchmarks, to near real-time streaming settings, such as interactive mobile robots, necessitates a methodology for appropriately feeding the HAR model with input data. To address this, we employ an overlapping sliding window technique that comprises a set of $n$ frames sampled at a rate of $sr$. The duration of the sliding window ($t_{sw}$) is calculated as $t_{sw} = n/sr$. Additionally, we define a prediction window ($t_{pw}$) as the time between two subsequent sliding windows, calculated as $t_{pw} = t_{sw} - (n-m)/sr$, where $m$ represents the number of newly introduced frames. Due to the overlapping nature of consecutive windows, multiple predictions may arise for a given time period, originating from different sliding windows. In Figure \ref{fig:har_sliding_window}, for instance, the first sliding window with duration $t_{sw}$ spans three distinct prediction windows, each with a duration of $t_{pw}$. Upon making a prediction, we store it and await the arrival of all predictions associated with that time period. As depicted in Figure \ref{fig:har_sliding_window}, the third time period encompasses three separate predictions from the second, third, and fourth sliding windows.

In the process of obtaining the final prediction for a specific prediction period, it is necessary to combine multiple individual predictions. The straightforward approach to address this is through a majority voting mechanism. However, in practical scenarios, particularly when the user is undergoing a transition stage, the outcomes of sequential time windows tend to exhibit instability. To mitigate this challenge, we leverage the probabilities generated by the HAR model. For each prediction period, the probabilities from all contributing sliding windows are averaged. The final prediction is inferred by selecting the larger number above a certain threshold that has been fine-tuned through empirical experimentation.

\section{Experimental Results} \label{results}

For the training of the HAR models we used a server-grade system with two NVIDIA RTX3090 GPUs (24GB). For the experimental evaluation we utilised an OZZIE mobile robot\footnote{\href{https://www.ozzie-robotics.com}{OZZIE robotics (https://www.ozzie-robotics.com)}} with the following specs: Intel i5 CPU, NVIDIA GTX1650 GPU (4GB) and Orbbec Astra RGBD camera. The pre-processing steps regarding the preparation of the sliding windows, as described in Section \ref{subsec:har}, along with the post-processing of the results were executed on the CPU of the robot, while the HAR models were deployed on the GPU, with the relevant metrics being discussed in Sections \ref{subsec:har_comparison} and \ref{subsec:mobile_deployment} and depicted in Figure \ref{fig:mem_inftime}. Lastly, the system is built upon and utilises the following libraries: python 3.10, torch 1.13.1, CUDA 11.6, CUDNN 8.8.1 and MMAction2 \cite{2020mmaction2}.

\subsection{Dataset Description} \label{subsec:dataset}

\begin{table*}[!hbt]
\centering
\captionof{table}{Dataset statistics}\label{dataset_statistics}
\begin{tabular}{ccccccccc}
\hline
Class & \begin{tabular}[c]{@{}c@{}}Videos\\ \#\end{tabular} & \begin{tabular}[c]{@{}c@{}}Avg Duration\\ (s)\end{tabular} & \begin{tabular}[c]{@{}c@{}}Ours\\ (\%)\end{tabular} & \begin{tabular}[c]{@{}c@{}}NTU\\ (\%)\end{tabular} & \begin{tabular}[c]{@{}c@{}}MSR\\ (\%)\end{tabular} & \begin{tabular}[c]{@{}c@{}}UTD\_MHAD\\ (\%)\end{tabular} & \begin{tabular}[c]{@{}c@{}}Weizmann\\ (\%)\end{tabular} & \begin{tabular}[c]{@{}c@{}}ISLDAS\\ (\%)\end{tabular} \\
\hline  \hline
drinking          & 238 & 4.07  & 124(52)   & 94(39)  & 20(8)    & 0(0)   & 0(0)  & 0(0)\\
eating            & 271 & 4.65  & 157(58)   & 94(35)  & 20(7)    & 0(0)   & 0(0)  & 0(0)\\
sitting           & 227 & 3.43  & 61(27)    & 94(41)  & 40(18)   & 32(14) & 0(0)  & 0(0)\\
standing          & 206 & 3.02  & 60(29)    & 94(46)  & 20(10)   & 32(16) & 0(0)  & 0(0)\\
walking           & 352 & 2.73  & 96(27)    & 0(0)    & 20(6)    & 0(0)   & 10(3) & 226(64)\\
lying             & 210 & 3.33  & 96(46)    & 94(45)  & 20(10)   & 0(0)   & 0(0)  & 0(0)\\
talking on phone  & 207 & 4.58  & 93(45)    & 94(45)  & 20(10)   & 0(0)   & 0(0)  & 0(0)\\
\hline  \hline
Total             & 1711 & 3.68 & 687(40.15) & 564(32.96) & 160(9.35) & 64(3.74) & 10(0.58) & 226(13.2)\\
\hline
\end{tabular}
\end{table*}

While there are datasets such as MSRDailyActivity3D \cite{6247813} that contain a diverse range of human actions, they fall short of capturing the full set of activities that we are attempting to identify. For instance, NTU RGB-D \cite{shahroudy2016ntu} lacks actions such as walking and lying, whereas UTD-MHAD \cite{7350781} only includes standing and sitting actions. NTU RGB-D videos were recorded from a different angle than the robot's view, and MSRDailyActivity3D, while covering a wider range of activities, has only twenty samples per class, which is insufficient for training our deep learning models. To address these limitations, we've created a new dataset with seven categories: drinking, eating, sitting, standing, walking, lying down, and talking on the phone. The dataset was created by recording RGB videos of 10 participants interacting with the robot, utilising its RGB camera in diverse setups. These setups included different lighting conditions, occlusions, scales, and varying positions of the robot relative to the users. By capturing such varied interactions, the dataset aims to provide a comprehensive and realistic representation of real-world scenarios. This ensures that the AI model trained on this data can effectively handle different environmental challenges, recognise objects and actions even in partially obscured situations, and adapt to various user distances and angles. The dataset's richness in diversity enables the AI model to be more robust and capable of performing well in practical applications involving human-robot interactions.

As shown in Table \ref{dataset_statistics}, to further enhance and strengthen our dataset, we also incorporated additional samples from other publicly accessible datasets. We adopted a three train/test split experimental approach, following the UCF101 \cite{soomro2012ucf101} and HMDB51 \cite{6126543} dataset division procedure. For every action category in our dataset, we chose $80\%$ of the examples for training data and $20\%$ for testing data. To assess the overall performance of each one of the targeted models on our dataset, we averaged their scores over the three splits.

\subsection{Comparison of HAR Models} \label{subsec:har_comparison}

Our pipeline system's final stage, human action recognition, requires significant computing and memory resources. Thus, we conducted an ablation study to find the best HAR model for our system from a variety of state-of-the-art models, paying specific emphasis to edge deployability. We evaluated both 3D CNN models like C3D, X3D \cite{Feichtenhofer2020}, Slowonly \cite{Feichtenhofer2019a}, and R(2+1)D \cite{8578773}, as well as newly introduced video Transformers such as TimeSformer \cite{Bertasius2021IsSA}, and Swin-Transformer \cite{liu2021swin}. Each model was fine-tuned on our introduced dataset after being pre-trained on Kinetics-400 \cite{Kay2017TheKH}. 

\begin{table*}[!hbt]
\centering
\caption{State-of-Art HAR models comparison on the proposed dataset}
\label{tab:dataset_results}
\begin{threeparttable}
\begin{tabular}{cccccc}
\hline
Model               & \begin{tabular}[c]{@{}c@{}}Top1\\ (\%)\end{tabular}        & \begin{tabular}[c]{@{}c@{}}Params\\ (M)\end{tabular}     & \begin{tabular}[c]{@{}c@{}}FLOPS\\ (G)\end{tabular}         & \begin{tabular}[c]{@{}c@{}}Views\\ $(clips \times crops)$\end{tabular}    & \begin{tabular}[c]{@{}c@{}}Input Shape\\ $(F \times H \times W)$\end{tabular}         \\ 
\hline \hline
C3D                 & 88.53          & 78.02              & 38.54          & $10\times1$              & $16\times112\times112$      \\
R(2+1)D             & \textbf{98.82} & 63.55              & 53.14          & $10\times3$              & $8\times256\times256$      \\
Slowonly            & 98.53          & 31.64              & 54.75          & $10\times3$              & $8\times256\times256$      \\
TimeSformer         & 93.53          & 121                & 196            & $1\times3$               & $8\times224\times224$      \\
Swin-Transformer    & 96.47          & 49.5               & 166            & $4\times3$               & $32\times224\times224$      \\
X3D                 & 98.24          & \textbf{2.99}      & \textbf{6.39}  & $10\times3$              & $16\times256\times256$      \\
\end{tabular}
\begin{tablenotes}
	\item The variation in the input's spatial and temporal dimensions is due to the differences in the original works, where various proposed input shapes exist.
\end{tablenotes}
\end{threeparttable}%
\end{table*}

\begin{figure}[!hbt]
  \centering
  \begin{subfigure}[b]{0.8\linewidth}
    \includegraphics[width=\linewidth]{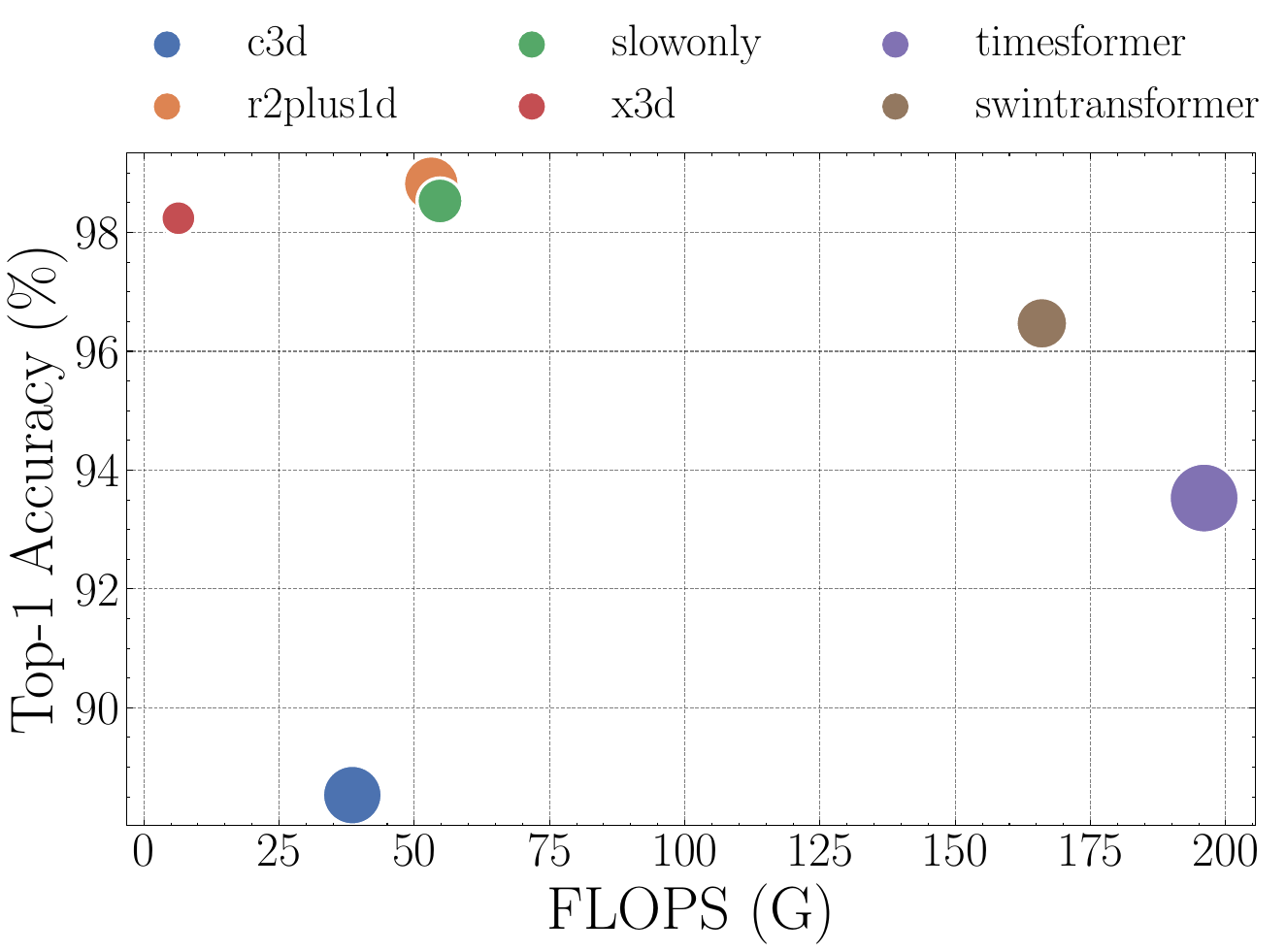}
    \caption{Top-1 Accuracy (\%) over FLOPS (G) evaluated on the proposed dataset.\\}
    \label{fig:top1_gflops}
  \end{subfigure}
  \hfill
  \begin{subfigure}[b]{0.8\linewidth}
    \includegraphics[width=\linewidth]{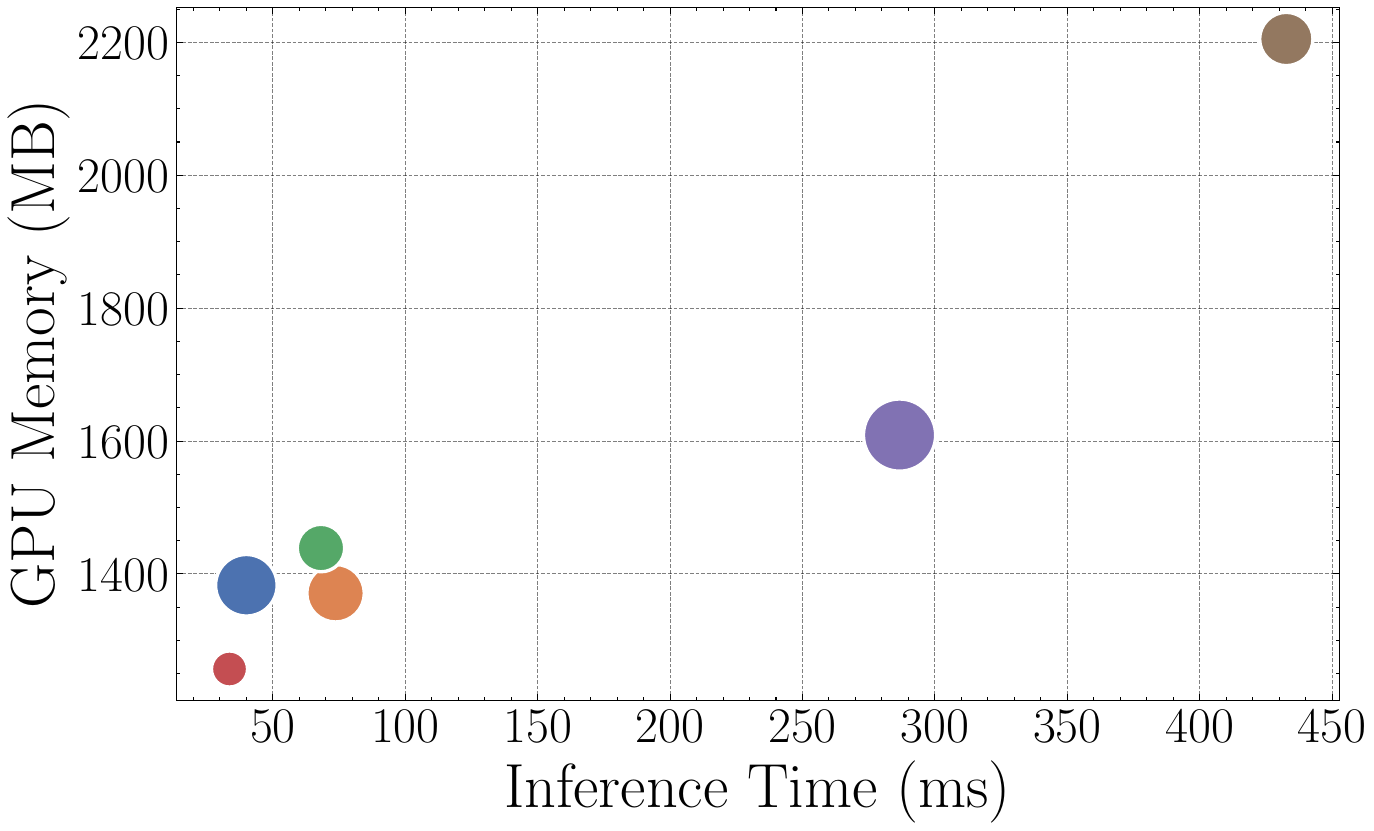}
    \caption{Execution speed (ms) over GPU memory consumption (MB) deployed on the mobile robot (GTX 1650).}
    \label{fig:mem_inftime}
  \end{subfigure}
  \caption{Comparison of state-of-art HAR models on both prediction accuracy and execution performance. The dots in both graphs represent the scale of the models' number of parameters, which ranges from 2.99 to 121 million. The larger the dot, the greater the number of parameters in the corresponding model.}
  \label{fig:models_comparison}
\end{figure}

To evaluate the models, we employed different pre-processing and sampling strategies based on the model's original configuration, as presented in Table \ref{tab:dataset_results}. The table provides details on the number of temporal clips and spatial crops used for evaluation, along with the Top1 accuracy and specific characteristics such as the number of parameters, FLOPS, and input shapes. As indicated in Table \ref{tab:dataset_results} and Figure \ref{fig:top1_gflops}, the X3D \cite{Feichtenhofer2020}, Slowonly \cite{Feichtenhofer2019a}, and R(2+1)D \cite{8578773} models achieved the highest Top1 accuracy, making them suitable choices considering only the recognition performance. Although the transformer models generally outperform the 3D CNN models in larger benchmarks like Kinetics-400, they were unable to surpass the performance of the 3D CNN models in our case due to the fact that Transformer models require a significantly larger amount of training data to achieve state-of-the-art performance, something that our mid-scale dataset could not provide.

\subsection{Deployment and Performance on the Mobile Robot} \label{subsec:mobile_deployment}

\begin{figure}[!hbt]
    \centering
    \includegraphics[width=0.95\linewidth]{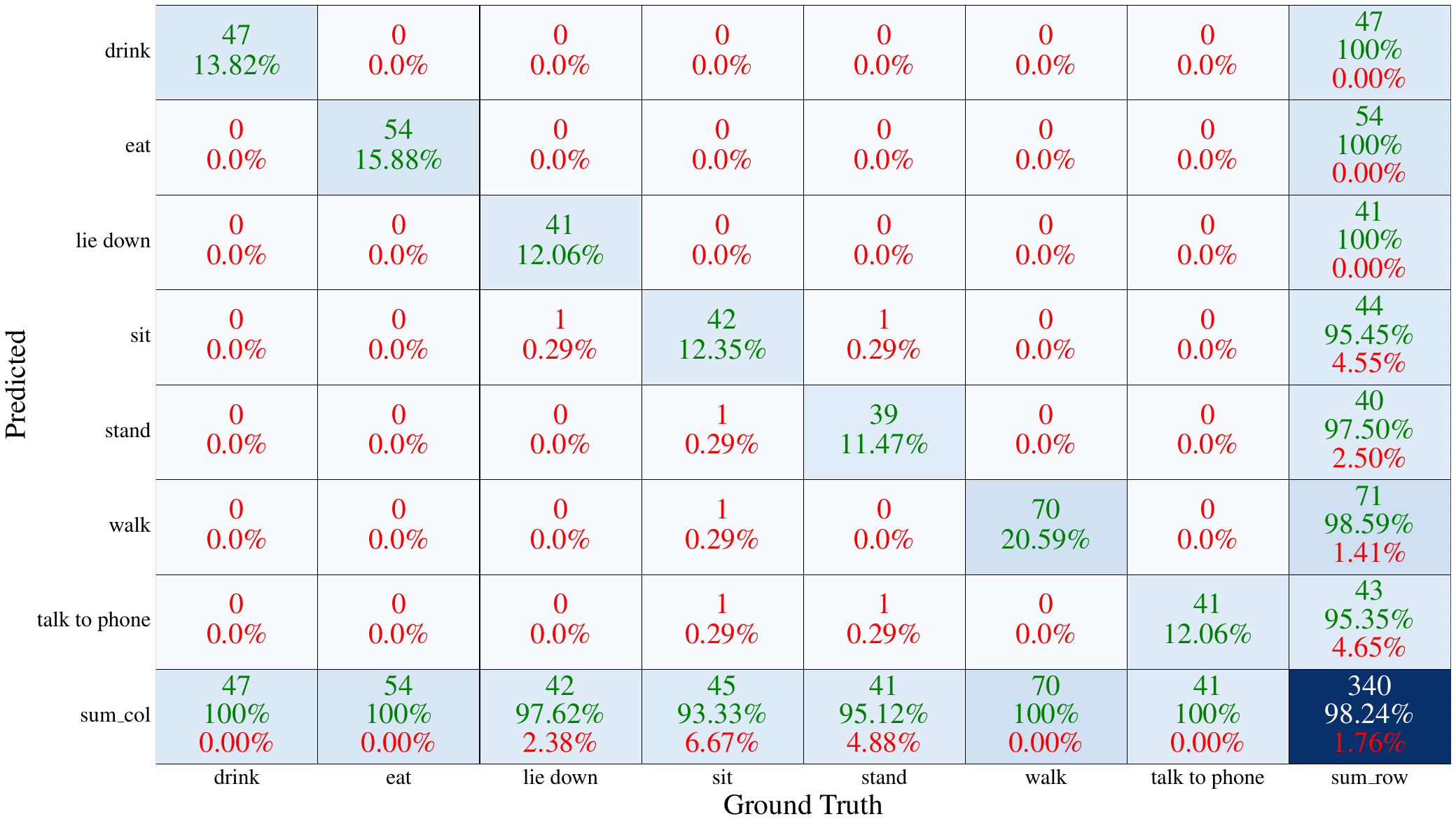}
    \caption{The Confusion Matrix of the X3D model on the proposed dataset}
    \label{fig:confusion_matrix}
\end{figure}

When deploying on edge platforms, like mobile robots, model prediction performance is not the only factor to consider. In such situations, near real-time performance and efficient resource utilisation become crucial factors in determining the most appropriate model. To evaluate the models in our scenario, we deployed and evaluated each one on the OZZIE mobile robot, measuring the GPU memory consumption and inference time. The results of these evaluations are illustrated in Figure \ref{fig:mem_inftime}. Figure \ref{fig:models_comparison} demonstrates that the X3D model achieves the best balance between memory consumption and inference time, while maintaining its high prediction accuracy. The X3D model may operate at 29 FPS on the robot, whereas the pipeline can run at 8 FPS when all steps are activated. The primary factor contributing to the performance degradation in our system is the communication overhead and resource sharing between different stages of the process. In the context of the ROS, which utilizes the TCPROS communication protocol, these communication patterns play a critical role in facilitating data exchange and coordination among various components of the robotic system. Finally, Figure \ref{fig:confusion_matrix} provides a more in-depth analysis of the X3D model's performance on our proposed dataset depicting the respective confusion matrix. Due to variations in system hardware, DNN models employed across pipeline stages, and the proposed end-to-end solutions, it is difficult to directly compare our suggested method to the rest of the works previously describen in Section \ref{sec:background}.

\section{Conclusion} \label{conclusions}

In this study, we have introduced a Human Action Recognition pipeline designed specifically for autonomous mobile service robots. We prioritise lightweight and resource-efficient solutions at each stage of the pipeline to overcome mobile robot computational limitations and achieve near real-time, on-device performance with high FPS. To ensure the applicability of our HAR model to daily activities in domestic environments, we have recorded a new dataset from the robot's perspective and evaluated our system on that. In the future, we plan to explore the inclusion of additional semantic information such as skeletal keypoints or scene objects to enhance the system's performance. Future considerations might include adding activity classes to the dataset, and investigating ways to perform HAR while the mobile robot is moving to address camera motion issues.

\section*{Acknowledgement}
The research work was supported by the Hellenic Foundation for
Research and Innovation (H.F.R.I.) under the “First Call for H.F.R.I.
Research Projects to support Faculty members and Researchers and
the procurement of high-cost research equipment grant” (Project
Name: ACTIVE, Project Number: HFRI-FM17-2271

\bibliographystyle{ieeetr}
\bibliography{references}

\end{document}